\documentclass[conference]{IEEEtran}
\usepackage{graphicx}
\usepackage[caption=false]{subfig}
\usepackage{multicol}
\usepackage{booktabs}
\usepackage{adjustbox}

\ifCLASSINFOpdf
\else
\fi
\usepackage{algorithm}
\usepackage{algorithmic}
\begin{document}
\title{Using Neural Networks for Programming by Demonstration}

\author{\IEEEauthorblockN{Karan K. Budhraja, Hang Gao and Tim Oates}
\IEEEauthorblockA{Computer Science and Electrical Engineering\\
University of Maryland, Baltimore County\\
Baltimore, Maryland 21250, USA\\
Email: \{karanb1,hanggao1\}@umbc.edu, oates@cs.umbc.edu}}

\maketitle

\begin{abstract}
Agent-based modeling is a paradigm of modeling dynamic systems of interacting agents that are individually governed by specified behavioral rules. Training a model of such agents to produce an emergent behavior by specification of the emergent (as opposed to agent) behavior is easier from a demonstration perspective. Without the involvement of manual behavior specification via code or reliance on a defined taxonomy of possible behaviors, the demonstrator specifies the desired emergent behavior of the system over time, and retrieves agent-level parameters required to execute that motion. A low time-complexity and data requirement favoring framework for reproducing emergent behavior, given an abstract demonstration, is discussed in \cite{budhraja2016controlling,budhraja2019dissertation}. The existing framework does, however, observe an inherent limitation in scalability because of an exponentially growing search space (with the number of agent-level parameters). Our work addresses this limitation by pursuing a more scalable architecture with the use of neural networks. While the (proof-of-concept) architecture is not suitable for many evaluated domains because of its lack of representational capacity for that domain, it is more suitable than existing work for larger datasets for the Civil Violence agent-based model.
\end{abstract}

%\keywords{optimization, agent based model, transition, demonstration}

% \fi
\IEEEpeerreviewmaketitle

\section{Introduction}
\label{section:introduction}

\begin{figure}[!t]
\centering
\subfloat[]{\includegraphics[width=0.16\textwidth]{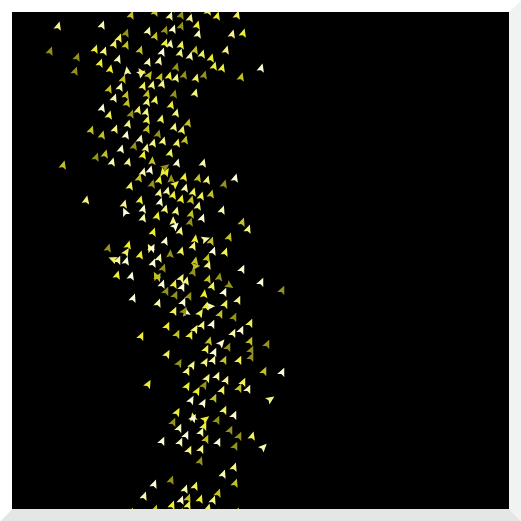}}
\subfloat[]{\includegraphics[width=0.16\textwidth]{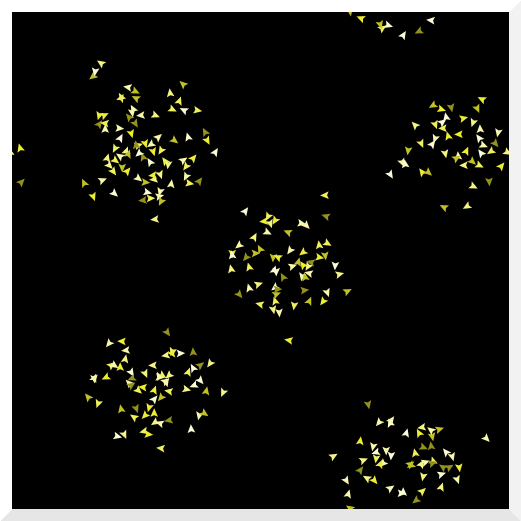}}
\subfloat[]{\includegraphics[width=0.16\textwidth]{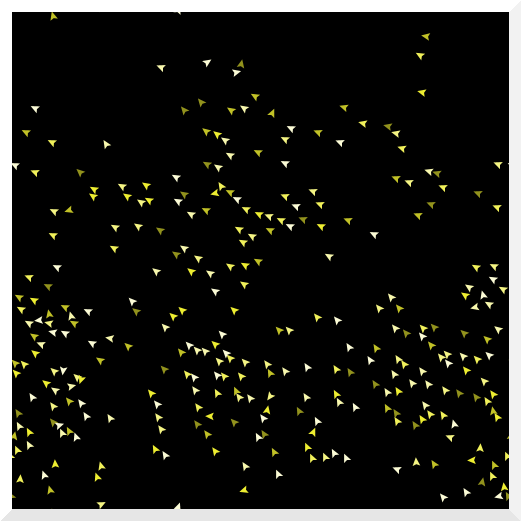}}
\caption{Different behaviors (SLPs) observed with the variation of the ALPs for the NetLogo-based simulation \cite{wilensky1999netlogo} for the flocking ABM \cite{wilensky1998netlogo,reynolds1987flocks}. These are also examples of different behavior demonstrations \cite{budhraja2016controlling}.}
\label{fig:varied_SLPs}
\end{figure}

An Agent-Based Model (ABM) is a computational model simulating interacting agent behavior through agent-level behavioral rule specification. Through interactions, the behaviors of individual agents produce more complex emergent collective behavior. Examples of ABMs include motion of humans in a crowd, spreading of diseases, and motion of groups of animals. In the first example, the ABM may specify average speed and direction of motion for each human, based on other humans around it. Simulation of such a system may then lead to behaviors such as formation of groups with aligned motion. The individual behavior of an agent is governed by control parameters called \textit{Agent-Level Parameters} (ALPs) for the purpose of this work. Different values of ALPs lead to different emergent behaviors in the ABM. Values quantifying such emergent behavior are called \textit{Swarm-Level Parameters} (SLPs). Variation of the three ALP values results in the visualization of different behaviors. An example of such variance is shown in Figure \ref{fig:varied_SLPs}.

%Figure \ref{fig:netlogoui} provides an example of an ABM simulation.

\begin{figure}
\centering
\includegraphics[width=0.45\textwidth]{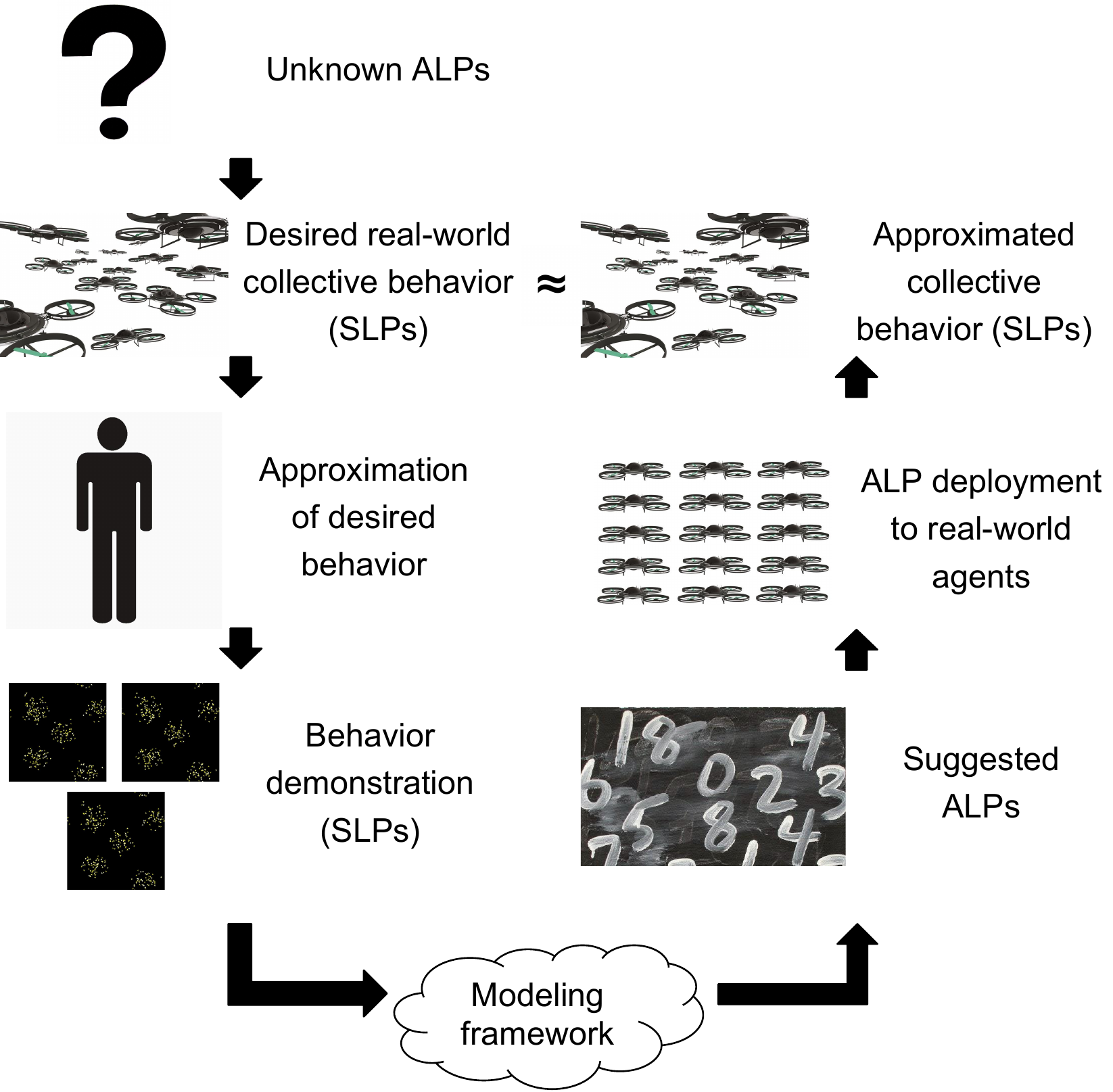}
\caption{An overview of this work being applied to an example. ALPs estimated for a demonstration can be deployed to a swarm of agents to replicate the demonstrated behavior. Impact of the ALPs is dependent on the ABM governing the swarm. However, the modeling framework is independent of the ABM used (since ABMs provide corresponding ALPs). The system uses an ABM as a black box to which ALPs are provided as input and SLPs are observed as output \cite{budhraja2016controlling}.}
\label{fig:overview}
\end{figure}

When demonstrating swarm behavior, it is easier for demonstrators to specify SLPs than it is to specify ALPs. Following from \cite{budhraja2016controlling,budhraja2017controlling} (the AMF$^{+}$ framework), we assume the demonstrators to be more tactically, rather than technically, skilled. This means that they can specify desired collective behavior but are not required to translate it to mathematical form. Given an abstract behavior specification (such as shape), a model is required to interpret the parameters, and estimate the ALPs needed to produce it. Figure \ref{fig:overview} depicts a procedural outline of how such a system is applied for this purpose. The task is to replicate observed or desired real-world swarm behavior. A demonstrator describes the behavior using images or a video of agent motion and uses our framework to estimate associated ALPs. These parameters can then be deployed to real-world agents to manifest the required behavior. Work in \cite{budhraja2016controlling,budhraja2017controlling} addresses a level of abstraction as in visual demonstration with a lower time complexity than stochastic search algorithms and without depending on prior knowledge about probable collective behaviors to follow demonstrations. 

%\begin{figure}[t] 
%\centering 
%\includegraphics[width=0.45\textwidth]{netlogoui.png} 
%\caption{Sliders in this NetLogo-based ABM simulation \cite{wilensky1999netlogo} define ALP values for each agent, such as alignment and separation with respect to other agents. With high alignment and low separation, agents begin to move in groups over time. The number of such groups or average size of groups are then examples of SLPs \cite{miner2010dissertation}.} 
%\label{fig:netlogoui} 
%\end{figure} 

%shown in Figure \ref{fig:netlogoui}

While work in \cite{budhraja2016controlling,budhraja2017controlling,budhraja2019dissertation} describes and builds on the AMF$^{+}$ framework, the authors acknowledge and adhere to its limitation in scalability. The search space used to map from SLPs to ALPs grows exponentially with the number of ALPs \cite{miner2010dissertation}. Our work therefore overlooks the low time complexity and data requirements of the AMF$^{+}$ framework and pursues a more scalable architecture with the use of neural networks.

Section \ref{section:relatedWork} describes existing work in the context of our contribution. The problem of using neural networks for the purpose of programming ABMs by demonstration is formalized in Section \ref{section:usingNeuralNetworks:problemDefinition} and is followed by the proposed solution in Section \ref{section:usingNeuralNetworks:proposedMethod}. Experimental evaluations in Section \ref{section:usingNeuralNetworks:experiments} are then used to evaluate the proposed neural network architecture and compare with the AMF$^{+}$ framework. At the cost of increased data requirements, the proposed neural network architecture shows potential to compete in performance with the AMF$^{+}$ framework and serves as a scalable alternative. It is also observed that using contacted networks facilitates better learning of the mapping between SLPs and ALPs than the direct use of a single network.

\section{Related Work}
\label{section:relatedWork}

Swarm behavior may be encoded as a \textit{fitness function} \cite{d2010evolving,reeder2015team,pickem2016safe} or using a high-level programming language \cite{pianini2015protelis,brambilla2012property,brambilla2015property,massink2013use}). 
Work in \cite{brambilla2013swarm} enumerates common behaviors and work in \cite{francesca2014automode} uses pre-existing behavior modules to describe emergent swarm behavior. These models depend on prior specification of possible behavior. Hierarchical training methods \cite{sullivan2012learning,freelan2014towards} require that the demonstrator manually decompose the task into a hierarchy of subtasks. Our work does not impose such a requirement on the demonstrator. Work in \cite{lancaster2015predicting} using probabilistic graphs to encode spatial distributions of agents requires them to be hand-made. Emergent behavior is formulated as an inverse reinforcement learning problem in \cite{vsovsic2016inverse}, but imposes the creation of controllers per agent with specification of a space of \textit{states}, \textit{observations} and \textit{actions}, among other MDP characteristics. However, our work addresses a demonstrator with more abstract requirements for the user by removing the requirement for technical specification of the demonstration. 

Work in \cite{miner2009rule,miner2008learning,miner2008swarm} discusses the use of abstraction of swarm behavior based on specified rules that relate ALPs and SLPs. However, it requires the manual specification of these rules. Subsequent work in \cite{miner2009learning,miner2009predicting,miner2010dissertation,miner2010github} then establishes a framework to produce functional mappings between ALPs and SLPs. This framework is called the ABM Meta-Modeling Framework (AMF). The (many-to-one) mapping from ALPs to SLPs is called Forward Mapping (FM). The (one-to-many) mapping from SLPs to ALPs is called Reverse Mapping (RM). The AMF framework generates FM and RM by sampling from training data, consisting of observed correspondence pairs between ALPs and SLPs. Work in \cite{budhraja2016controlling} (the AMF$^{+}$ framework) extends \cite{miner2009learning,miner2009predicting,miner2010dissertation,miner2010github} by using the RM to suggest a single configuration of ALPs that may have produced a given demonstration. The many points (in ALP space) returned on querying RM are condensed to a single representative point.

Following evaluated methods in \cite{budhraja2019dissertation}, research in the domain of inverse kinematics is identified to facilitate the application of neural networks to generate one-to-many mappings. Work using hierarchical neural networks \cite{wu2016set} uses a hierarchy of neural networks to generate input variables independently. This does, however, require manual identification of the number of possible sets in the behavioral distribution with respect to the input variable. Other work \cite{daya2010applying} also uses multiple neural networks for this problem, but in a different manner. The networks are trained and optimized in parallel as parts of a larger ensemble network. This approach also requires manual specification in the form of the number of sub-networks used.

Another approach \cite{nagataa2016neural} describes a method to circumvent susceptibility to plateaus when optimizing neural network loss. A method of alternating between training on the full dataset and a subset is proposed, where the subset consists of data points corresponding to maximum error. While this iterative approach is useful to continue learning for extended epochs, it may be unnecessary for the current exploratory nature of this work. Alternatively, approaches may remove instances of duplicate behavior to convert the one-to-many mapping represented by the training data to a one-to-one mapping \cite{duka2014neural}. Such discarding of information may cause a loss in abstract relationship representations between input and output. Some approaches employ shallow neural networks to produce a direct mapping from SLPs to ALPs \cite{srisuk2017inverse,jha2014neural}. These methods do not make use of the one-to-many nature of the mapping and may lead to reduced quality of inference (during neural network optimization) if the source ALPs for the given SLPs vary significantly (e.g., if they belong to distant clusters in ALP space).

The method proposed in recent feedback-based work \cite{almusawi2016new} is exemplified using a robot arm. It uses the retried ALPs (joint angles) and applies them to the driving unit (ABM). The observed position and orientation of the robot arm (SLPs) are then provided as input to the neural network in the form of control signals. This control method is based on feedback and is therefore not applicable to our primary problem. Such control may, however, be useful as a parallel for feedback to the framework \cite{budhraja2018implementing}. An important observation in this work is the combination of RM and FM processes in some form. 

Prior work \cite{jordan1992forward} describes the use of modular distal teacher network to solve the inverse kinematics mapping problem. While that work additionally discusses many control-based approaches for temporal data, the method proposed in this work is similar to the feedforward network proposed in prior work \cite{jordan1992forward}. The FM and RM networks proposed in Section \ref{section:usingNeuralNetworks:proposedMethod} are symmetrically opposite in structure, however, unlike their counterparts \cite{jordan1992forward}.

Also note that all these works that serve to map from SLPs to ALPs are trained with the intuition of producing neural networks with an optimal one-to-one mapping that serves as a proxy for the underlying one-to-many mapping. This means that the networks are able to learn decision boundaries that best separate similar behaviors (except in the case of bypassing one-to-many mapping \cite{duka2014neural}).

\section{Problem Definition}
\label{section:usingNeuralNetworks:problemDefinition}

Given a dataset of ALP and SLP correspondences, the current solution described in \cite{budhraja2016controlling,budhraja2017controlling,budhraja2019dissertation} is observed to be limited in scalability with respect to ALPs. The problem, then, is to propose a more architecture that may learn the RM mapping (SLPs to ALPs) as an alternative to the AMF$^{+}$ framework. This requires the formulation of a suitable neural network architecture.

\section{Proposed Method}
\label{section:usingNeuralNetworks:proposedMethod}

This section proceeds to describe two proposed architectures to solve this problem.

\subsection{Proposed MLP Architecture}
\label{subsection:usingNeuralNetworks:proposedMlpArchitecture}

Similar to methods proposed in existing work \cite{miner2010dissertation,almusawi2016new,jordan1992forward}, the primary idea behind the proposed method is to allow the FM architecture to assist the RM architecture in learning an approximation of the required one-to-many mapping. This comprises of two stages.

The first stage of the proposed approach is therefore to train a neural network to learn the many-to-one FM mapping. Similar to prior work \cite{miner2010dissertation,jordan1992forward}, an FM regression model provides a surrogate for expensive ABM simulations. The second stage of the proposed approach is to formulate a network for RM mapping. Because related research in inverse kinematics demonstrates the effectiveness of MLP neural networks, the architecture employed for this layer is MLP, similar to FM. 

While the FM MLP network can be trained independently using MSE loss, training the RM MLP network independently with such loss is not as useful as it is for FM because of the one-to-many nature of mappings. For this reason, the FM network is then concatenated to the RM network to produce the prospective output behavior, given RM network outputs. This allows for observation of output behavior similar to the use of feedback \cite{almusawi2016new} but in a faster manner. The difference between the concatenated FM network outputs and RM network inputs can then be used as MSE for training loss. Note that the FM section of the network is not trained while training the RM section of the network. The output of the system to the user, however, would be the intermediate output of this concatenated network, i.e., the output of the RM network. For clarity, the concatenated network is called the FM$+$RM network.

For exploratory simplicity, both FM and RM architectures are identical in layer structure, but with opposite layer order from input to output. Layer sizes are adapted based on the dimensions of ALPs and SLPs, with the intuition that larger vectors of ALPs and SLPs would require more complex networks for effective learning of mappings. The number of nodes in hidden layers is assigned proportional to the number of ALPs or SLPs, or the square of that value. Examples of network architecture are shown in \cite{budhraja2019dissertation}. Note that all architectures discussed in this section are composed of fully connected layers.

The AMF$^{+}$ framework uses a small subset of the total training data available (which may also be selected at random if not using \textit{Dataset Selection} as in \cite{budhraja2017dataset}) to generate FM and RM. The neural networks discussed in this work, however, are trained using the full extent of the available training data. The specific sizes of subsets of training used by the AMF$^{+}$ framework are documented in \cite{budhraja2019dissertation}. The training data requirement of $430$ to $2700$ points is therefore changed to $9,000$ points (this is the size of training data per cross-validation fold). The loss used to train this architecture is reconstruction error (between inputs to and outputs from the FM$+$RM network) in the form of Mean Squared Error (MSE).

%\begin{figure*}[!t]
%\centering
%\subfloat[]{\includegraphics[width=0.33\textwidth]{nn/architecture/flocking/rm.png}}
%\subfloat[]{\includegraphics[width=0.33\textwidth]{nn/architecture/flocking/fm.png}}
%\subfloat[]{\includegraphics[width=0.33\textwidth]{nn/architecture/flocking/model.png}}
%
%\subfloat[]{\includegraphics[width=0.33\textwidth]{nn/architecture/civil_violence/rm.png}}
%\subfloat[]{\includegraphics[width=0.33\textwidth]{nn/architecture/civil_violence/fm.png}}
%\subfloat[]{\includegraphics[width=0.33\textwidth]{nn/architecture/civil_violence/model.png}}
%\caption{Proposed RM$+$FM concatenated MLP architectures exemplified for the Flocking ($(a)-(c)$) and Civil Violence ($(d)-(f)$) ABM domains. Networks from left to right correspond to RM, FM and RM$+$FM networks respectively. The number of hidden units in layers at the center of the FM and RM networks scale as the squared of the dimensionality of ALPs and SLPs. Other layers are scaled linearly. Architectures used for other domains are shown in Appendix \ref{app:usingNeuralNetworks}.}
%\label{fig:usingNeuralNetworks:architecture}
%\end{figure*}

\subsection{Proposed One-to-Many Architecture}

While the network described in Section \ref{subsection:usingNeuralNetworks:proposedMlpArchitecture} is a solution to approximate the one-to-many RM mapping, the output of a single query for a fixed randomness setting is unique. If a one-to-many mapping is mandated for the RM network, then probabilistic architectures such as variational encoding layers (or similar) may be introduced \cite{kingma2013auto,lucas2017auxiliary}. To produce multiple RM outputs in such a setting of fixed environment randomness, we therefore use variational inference. We assume that the points corresponding to the one-to-many mapping represented by the output of the RM network are a function of hidden variables. These hidden variables are assumed to adhere a Gaussian distribution and are produced as a function of the input SLPs. The hidden variables are therefore be represented using mean ($\mu$) and variance ($\sigma^{2}$) values. Specifically, nodes are encoded using ($\mu$, log($\sigma^{2}$)) pairs. For exploratory simplicity, the number of hidden variables used for variational inference is identical to the size of the layer at the middle of the network (the square of the number of SLPs).

The loss used to train this architecture consists of two parts. The first is 
the reconstruction error (between inputs to and outputs from the FM$+$RM network) in the form of MSE. The second is the Kullback–Leibler (KL) divergence \cite{kullback1951information} to promote separation of fitted Gaussian surfaces for the variational layer. The contribution of each loss to the total loss may be weighted using a parameter $\alpha$, corresponding to weights of ($1-\alpha$) and $\alpha$ for the first and second loss components respectively.

\begin{figure*}[!t]
\centering
\subfloat[]{\includegraphics[width=0.33\textwidth]{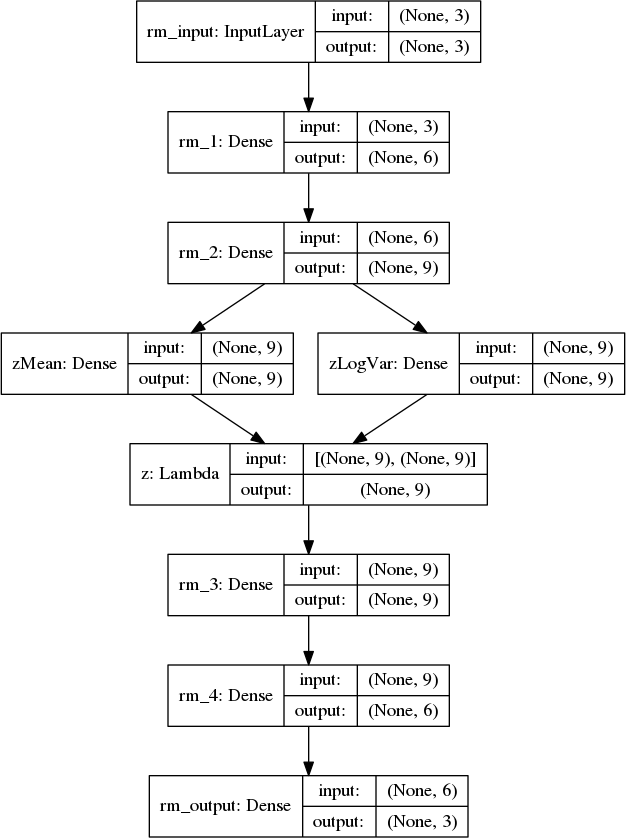}}
\subfloat[]{\includegraphics[width=0.33\textwidth, height=8.0cm, keepaspectratio]{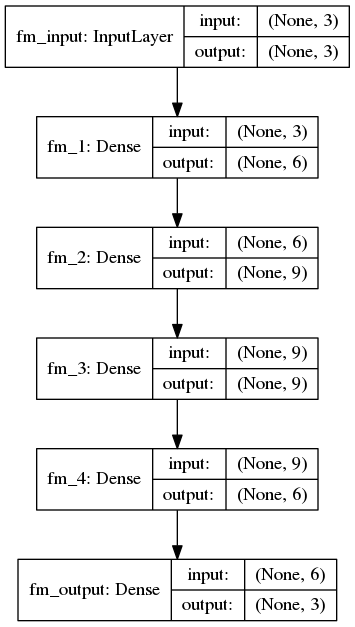}}
\subfloat[]{\includegraphics[width=0.25\textwidth, height=8.0cm, keepaspectratio]{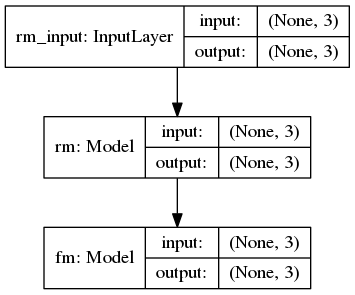}}
\caption{Proposed RM$+$FM concatenated variational architecture exemplified for the Civil Violence ABM domain. Networks from left to right correspond to RM, FM and RM$+$FM networks respectively. The number of hidden units in layers at the center of the FM and RM networks scale as the squared of the dimensionality of ALPs and SLPs. Other layers are scaled linearly.}
\label{fig:usingNeuralNetworks:architecture_civil_violence_variational}
\end{figure*}

\section{Experiments}
\label{section:usingNeuralNetworks:experiments}

The experiment data used for evaluation of the neural network alternatives to the framework is a subset of that used in \cite{budhraja2019dissertation}. While the $10$-fold cross-validation remains the same, the comparative performance of the AMF$^{+}$ framework used for reference is when using its current state (as in \cite{budhraja2019dissertation}) without outlier detection. Work in \cite{budhraja2019dissertation} establishes that the performance of the framework is least sensitive to this module. It is therefore unused in comparison, given the significant increase in time complexity caused by the use of outlier detection.

As is for all datasets used to train the AMF$^{+}$ framework, data is normalized to the range $[0,1]$. Additionally, standard scaling of input data to have a mean of $0$ and variance of $1$ is established as an effective method to tailor data for neural network training \cite{lecun2012efficient}. While such standardization of ALPs improves neural network performance, standardization of SLPs is observed to cause degradation. For this reason, SLPs are not standardized. Subsequently, favorable activation functions for the FM and RM network are empirically observed as sigmoid \cite{marreiros2008population} and ReLU \cite{nair2010rectified} respectively.

The ABMs used for evaluation follow from work in \cite{budhraja2019dissertation} and are summarized in Table \ref{tab:benchmarking:amf+Abms} as an excerpt from their descriptions in \cite{budhraja2019dissertation}.

\begin{table}[!t]
\centering
\caption{A summary of the number of ALPs and SLPs corresponding to various evaluated ABMs. Visual domains, for which SLPs are demonstrated by an image, are specified using ticks.}
\begin{tabular}{c *2c c}
\toprule
\toprule
ABM & $\#$ALPs & $\#$SLPs & Visual Domain \\
\midrule
Forest fire \cite{wilensky1997netlogo} & 1 & 1 & {} \\
Schelling \cite{schelling1971dynamic} & 2 & 1 & {} \\
Turbulence \cite{wilensky2003netlogo} & 2 & 1 & $\sqrt{}$ \\
EUM \cite{masad2016agents,de2002predicting} & 2 & 2 & {} \\
AIDS \cite{wilensky1997netlogoAIDS} & 2 & 2 & {} \\
Flocking \cite{wilensky1998netlogo,reynolds1987flocks} & 3 & 2 & $\sqrt{}$ \\
Civil violence \cite{epstein2002modeling} & 3 & 3 & {} \\
\bottomrule
\bottomrule
\end{tabular}
\label{tab:benchmarking:amf+Abms}
\end{table}

Equations referenced in this section are summarized as follows using their explanations in \cite{budhraja2019dissertation}. Work in \cite{brouwer2004feed} approximates Equation \ref{eq:brower_sin}, where $x \in [0, 1]$ and evaluates $f^{-1}(x)$. Work in \cite{wu2016set} describes a method to invert the forward kinematics equations shown in Equation \ref{eq:forward_kinematics_p1} and Equation \ref{eq:forward_kinematics_p2}, where $a,b \in [-2\pi, 2\pi]$. The work evaluates the computation of inputs \textit{a} and \textit{b} independently with datasets partitioned as $(a,p_1,p_2)$ and $(b,p_1,p_2)$. These two problem sets are called \textit{A} and \textit{B} respectively.

\begin{equation}
f(x) = x + 0.5*sin(2* \pi *x)
\label{eq:brower_sin}
\end{equation}

\begin{equation}
p_1 = cos(a+b) + sin(a)*sin(b)
\label{eq:forward_kinematics_p1}
\end{equation}

\begin{equation}
p_2 = sin(a+b) + cos(a)*sin(b)
\label{eq:forward_kinematics_p2}
\end{equation}

This section is divided into two subsections, following the structure of Section \ref{section:usingNeuralNetworks:proposedMethod}. Evaluations use 
\textit{Output Difference} \cite{budhraja2018implementing} and MSE as a measure of performance. \textit{Output Difference} is the Euclidean distance between the demonstration SLPs and the SLPs generated by framework suggestions.

\subsection{Proposed MLP Architecture}
\label{subsection:usingNeuralNetworks:experiments:proposedMlpArchitecture}

First, to demonstrate the viability of the RM MLP architecture, its performance is compared to several other architectures based on existing work. These networks focus on the problem of inverse kinematics, and are therefore evaluated on Problem set A \cite{wu2016set} data. Train and test data consist of a single fold and as in \cite{budhraja2019dissertation} and regression error is evaluated using MSE. Table \ref{tab:usingNeuralNetworks:rm_architectures} shows that MSE is similar (on average) when adapting to other existing MLP architectures. This work therefore proceeds with the proposed MLP architecture as in Section \ref{section:usingNeuralNetworks:proposedMethod}. The evaluated architectures corresponding to Table \ref{tab:usingNeuralNetworks:rm_architectures} are shown in \cite{budhraja2019dissertation}.

\begin{table}[!t]
\centering
\caption{Mean ($\mu$) and standard deviation ($\sigma$) for MSE computed for various RM MLP architectures for the domain defined by Problem set A. After the completion of training, MSE is evaluated on the train and test data for the given cross-validation split. See \cite{budhraja2019dissertation} for corresponding architectures evaluated.}
\begin{tabular}{c | c c | c c}
\toprule
\toprule
Source Architecture & \multicolumn{2}{c}{Train MSE} & \multicolumn{2}{c}{Test MSE $\mu$} \\
\cline{2-5}
{} & $\mu$ & $\sigma$ & $\mu$ & $\sigma$ \\
\midrule
Proposed MLP Architecture & 0.5000 & 0.4515 & 0.5090 & 0.4519 \\
\cite{daya2010applying} & 0.4752 & 0.4447 & 0.5201 & 0.4820 \\
\cite{duka2014neural} & 0.4787 & 0.4369 & 0.5232 & 0.4622 \\
\cite{jha2014neural} & 0.4984 & 0.4525 & 0.5114 & 0.4549 \\
\cite{almusawi2016new} & 0.5000 & 0.4515 & 0.5090 & 0.4520 \\
\cite{nagataa2016neural} & 0.4877 & 0.4467 & 0.5170 & 0.4642 \\
\bottomrule
\bottomrule
\end{tabular}
\label{tab:usingNeuralNetworks:rm_architectures}
\end{table}

Next, to demonstrate the limitations of the direct use of an RM network, the RM network is first trained on the various datasets for independent network evaluation. Regression error is then computed as MSE. These values serve as reference for network architecture performance. This data is then accompanied by FM regression MSE and subsequent RM$+$FM MLP network MSE observed when training the concatenated network. Experiments are summarized in Table \ref{tab:usingNeuralNetworks:neural_network_mse_mean_sigma}. Unlike testing in \cite{budhraja2016controlling,budhraja2017controlling}, evaluations in Table \ref{tab:usingNeuralNetworks:neural_network_mse_mean_sigma} use the entirety of each test set instead of limiting to $10$ test instances. This is done for each of $10$ folds i.e., the neural network is independently trained once on training data for a given fold. This leads to a total of $10$ values used for averaging, each representing the mean data for that fold. This is with the exception of the forward kinematics domain and derived problem sets A and B \cite{wu2016set}. As in \cite{budhraja2019dissertation}, these domains contain a single split of test and train data. This is the same setting used for the AMF$^{+}$ framework and is therefore used for fair comparison. Additionally, the domain corresponding to \cite{brouwer2004feed} is used with a total dataset size of $10,000$ similar to other evaluated ABMs (except forward kinematics). For experimental consistency, a seed value of $0$ is used to initialize randomization. Further, all neural networks are trained for $1000$ epochs.

\begin{table*}[!t]
\centering
\caption{Mean ($\mu$) and standard deviation ($\sigma$) for MSE computed for the RM network and FM network independently, and for the concatenated RM$+$FM MLP architecture across several domains. After the completion of training, MSE is evaluated on the train and test data for a given cross-validation split. Equation \ref{eq:brower_sin} corresponds to the simulation model used in \protect\cite{brouwer2004feed}. Equation \ref{eq:forward_kinematics_p1} and Equation \ref{eq:forward_kinematics_p2} correspond to the simulation model used in \protect\cite{wu2016set}.}
\begin{tabular}{c | c c c | c c c | c c c | c c c}
\toprule
\toprule
ABM & \multicolumn{3}{c}{Train MSE $\mu$} & \multicolumn{3}{c}{Train MSE $\sigma$} & \multicolumn{3}{c}{Test MSE $\mu$} & \multicolumn{3}{c}{Test MSE $\sigma$}\\
\cline{2-13}
{} & RM & FM & RM$+$FM & RM & FM & RM$+$FM & RM & FM & RM$+$FM & RM & FM & RM$+$FM \\
\midrule
Forest fire & 1.0000 & 0.2129 & 0.2129 & 0.0000 & 0.0002 & 0.0002 & 1.0004 & 0.2130 & 0.2130 & 0.0180 & 0.0026 & 0.0026 \\
Schelling & 2.0000 & 0.0982 & 0.0982 & 0.0000 & 0.0002 & 0.0002 & 2.0000 & 0.0982 & 0.0982 & 0.0427 & 0.0022 & 0.0022 \\
Turbulence & 2.0001 & 0.0028 & 0.0028 & 0.0001 & 0.0001 & 0.0001 & 2.0012 & 0.0028 & 0.0028 & 0.0354 & 0.0012 & 0.0012 \\
EUM & 0.6474 & 0.0081 & 0.0224 & 0.0392 & 0.0033 & 0.0162 & 0.6489 & 0.0082 & 0.0228 & 0.0431 & 0.0029 & 0.0168 \\
AIDS & 0.6122 & 0.0064 & \textbf{0.0001} & 0.0228 & 0.0001 & 0.0000 & 0.6155 & 0.0065 & \textbf{0.0001} & 0.0386 & 0.0007 & 0.0000 \\
Flocking & 1.1872 & 0.0138 & 0.0018 & 0.0039 & 0.0001 & 0.0008 & 1.1925 & 0.0141 & 0.0018 & 0.0208 & 0.0005 & 0.0008 \\
Civil violence & 0.1590 & \textbf{0.0004} & \textbf{0.0001} & 0.0041 & 0.0000 & 0.0000 & 0.1627 & \textbf{0.0004} & \textbf{0.0001} & 0.0040 & 0.0000 & 0.0000 \\
\midrule
Equation \ref{eq:brower_sin} & 1.0000 & 0.0495 & 0.0495 & 0.0000 & 0.0001 & 0.0001 & 1.0006 & 0.0496 & 0.0496 & 0.0238 & 0.0009 & 0.0009 \\
\midrule
Equations \ref{eq:forward_kinematics_p1}, \ref{eq:forward_kinematics_p2} & 0.9792 & 0.0530 & 0.0703 & 0.0000 & 0.0000 & 0.0000 & 1.0220 & 0.0510 & 0.0691 & 0.0000 & 0.0000 & 0.0000 \\
Problem set A & 0.5000 & 0.0703 & 0.0703 & 0.0000 & 0.0000 & 0.0000 & 0.5090 & 0.0694 & 0.0694 & 0.0000 & 0.0000 & 0.0000 \\
Problem set B & 0.5000 & 0.0703 & 0.0703 & 0.0000 & 0.0000 & 0.0000 & 0.5274 & 0.0690 & 0.0690 & 0.0000 & 0.0000 & 0.0000 \\
\bottomrule
\bottomrule
\end{tabular}
\label{tab:usingNeuralNetworks:neural_network_mse_mean_sigma}
\end{table*}

It is evident from Table \ref{tab:usingNeuralNetworks:neural_network_mse_mean_sigma} that the use of an FM network to assist the RM network results in a significant decrease in MSE. It is also evident that all the trained networks do not overfit because the ratio of train and test MSE is close to $1$. The order of MSE values vary with the domain, showing the varied capacity of the neural networks to learn for those domains. Apart from the complexity of the data, these variations are also attributed to the varied network architectures used. For example, the architecture used for the Civil Violence domain has significantly more hidden nodes than the architecture used for the Forest Fire domain (see \cite{budhraja2019dissertation}). Note that to be competitive with the performance of the AMF$^{+}$ framework (mean Euclidean distance of the order of $1e-2$ as in \cite{budhraja2018improved,budhraja2019dissertation}), the neural networks would require MSE that is approximately of the order of $1e-4$. This is because MSE is proportional to the square of Euclidean distance and the number of SLPs for the evaluated domains is small. As a result, the AIDS and Civil Violence domains show potential for the use of their architectures to replace the AMF$^{+}$ framework.

\begin{figure*}
\centering
\subfloat[]{\includegraphics[width=0.33\textwidth]{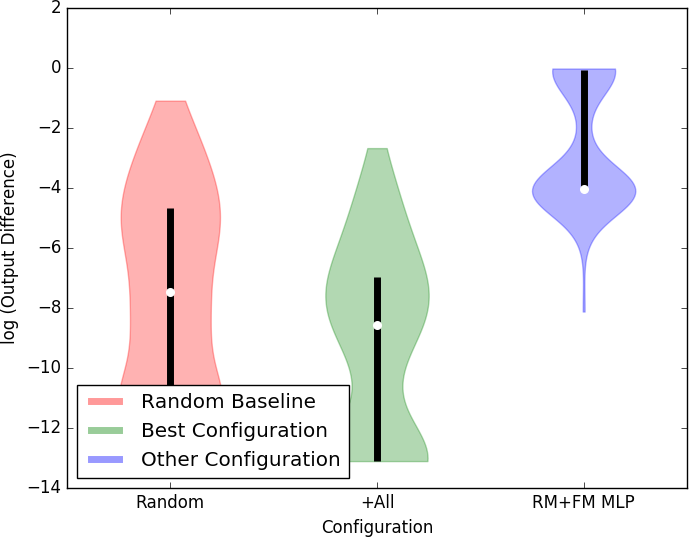}}
\subfloat[]{\includegraphics[width=0.33\textwidth]{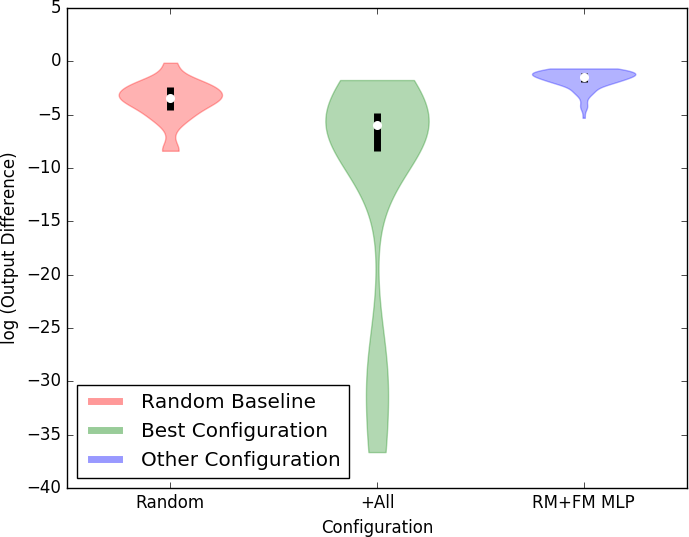}}
\subfloat[]{\includegraphics[width=0.33\textwidth]{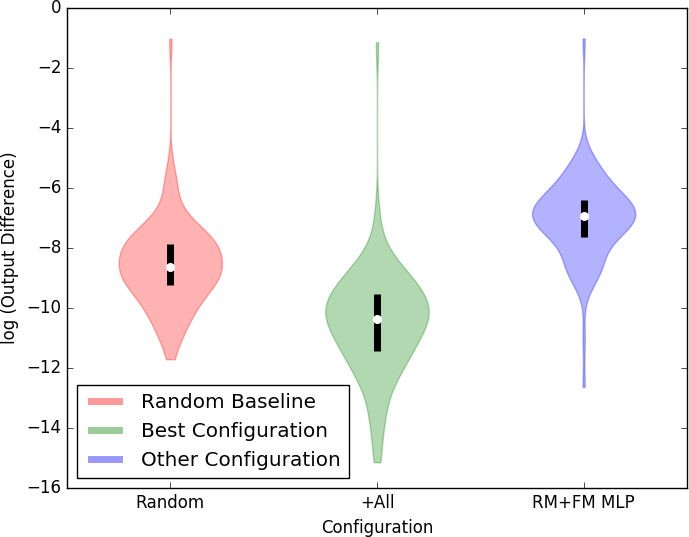}}

\subfloat[]{\includegraphics[width=0.33\textwidth]{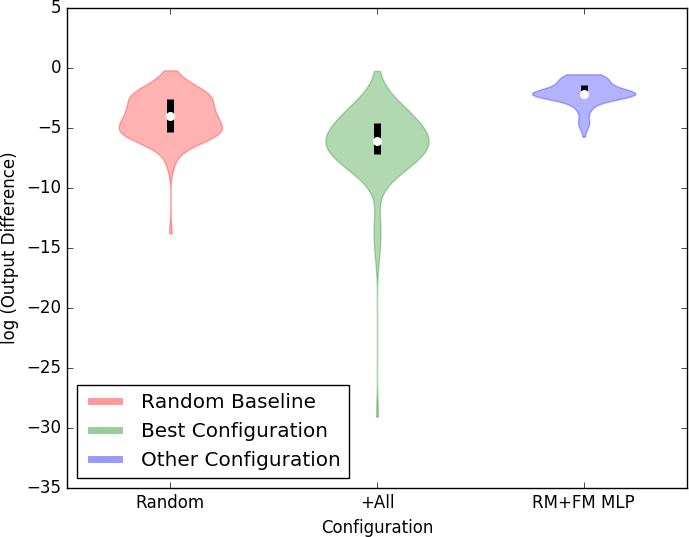}}
\subfloat[]{\includegraphics[width=0.33\textwidth]{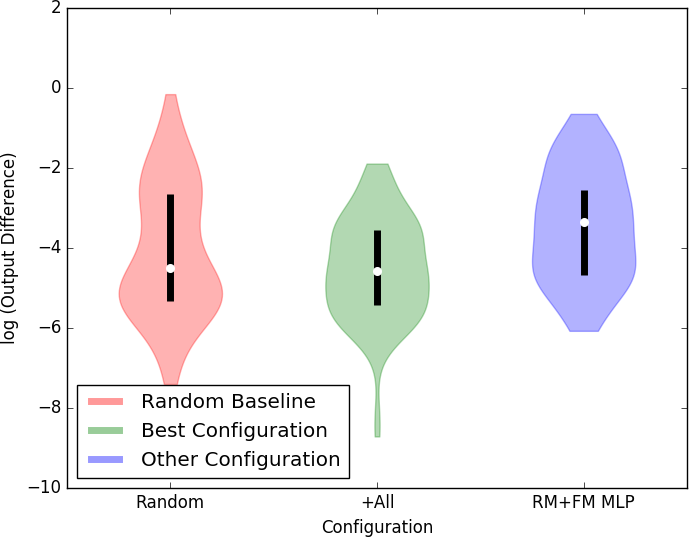}}
\subfloat[]{\includegraphics[width=0.33\textwidth]{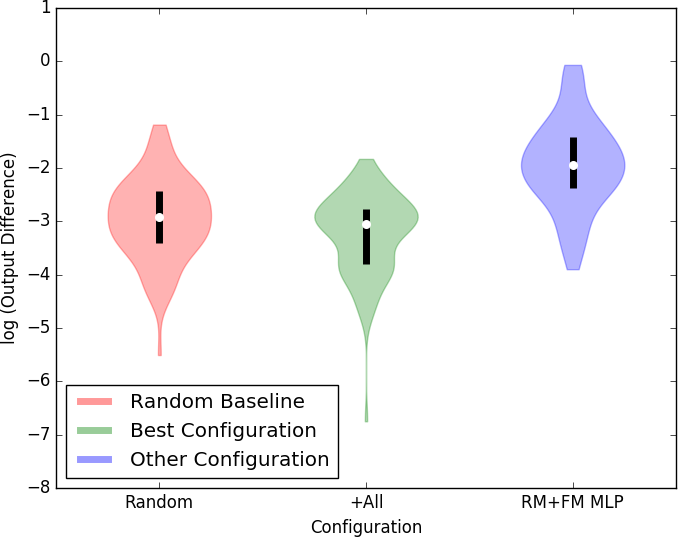}}

\subfloat[]{\includegraphics[width=0.33\textwidth]{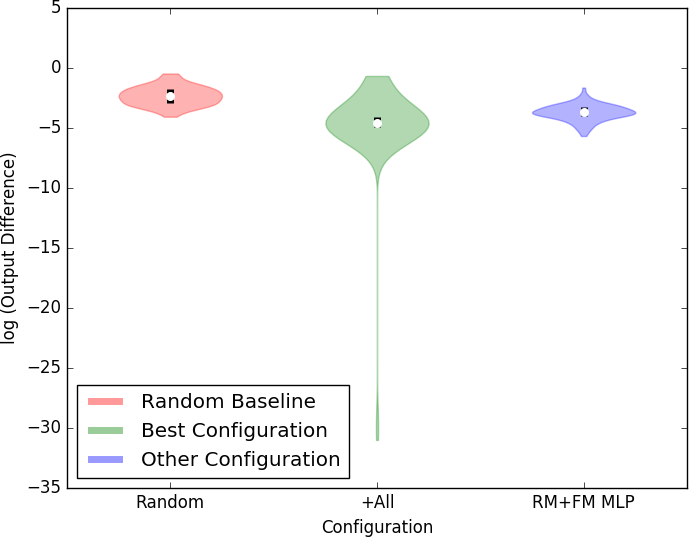}}
\caption{Comparison of the performance of the performance of using them RM$+$FM MLP network and the AMF$^{+}$ framework for various ABMs. ABMs used for Figures \textit{(a)} to \textit{(g)} are in correspondence to the top-to-bottom listing of ABMs in Table \ref{tab:benchmarking:amf+Abms}. The network performs better than the random baseline and becomes competitive with the AMF$^{+}$ framework for the Civil Violence domain (Figure \textit{(g)}).}
\label{fig:usingNeuralNetworks:comparison}
\end{figure*}

Next, the trained RM$+$FM MLP neural network therefore is evaluated for its suggested ALPs, i.e., the output at the end of the RM section of the network. The data used for these experiments is consistent with Table \ref{tab:usingNeuralNetworks:neural_network_mse_mean_sigma}. Similar to evaluations in \cite{budhraja2019dissertation}, the suggested ALPs are provided to the ABM simulator to produce corresponding SLPs. For comparison with the AMF$^{+}$ framework, error values are measured using Euclidean distance. Results for this evaluation are shown in Figure \ref{fig:usingNeuralNetworks:comparison}. With each test fold contributing $10$ test instances, data is averaged (median) over $100$ computed values (as in \cite{budhraja2019dissertation}).

For most ABM domains, the neural network shows significant deterioration in performance. This is expected given high train and test MSE values for the RM$+$FM network observed in Table \ref{tab:usingNeuralNetworks:neural_network_mse_mean_sigma}. We focus now on domains where the MSE values show potential for performance competitive with the AMF$^{+}$ framework: AIDS and Civil Violence. While the AIDS domain observes poor performance by the neural network, that observed for the Civil Violence domain is competitive with the performance of the AMF$^{+}$ framework. The median values of the logarithm of \textit{Output Difference} corresponding to the random baseline, the AMF$^{+}$ framework and the neural network for this domain are $-2.3567$, $-4.6212$ and $-3.7008$ respectively. The difference in performance for ALP suggestions for the AIDS and Civil Violence domains is caused by the difference in MSE for the FM network (see Table \ref{tab:usingNeuralNetworks:neural_network_mse_mean_sigma}). Even though the RM$+$FM network for the AIDS domain is able to reconstruct the SLPs from the output of the RM network, the high MSE of the FM network indicates that the mapping learned by the RM network does not output ALPs but an imprecise representation of them. Competitive performance of the RM$+$FM network for the Civil Violence domain demonstrates the potential effectiveness of the proposed neural network architecture when the network has sufficient capacity to learn data mappings.

\begin{figure}
\centering
\includegraphics[width=0.45\textwidth]{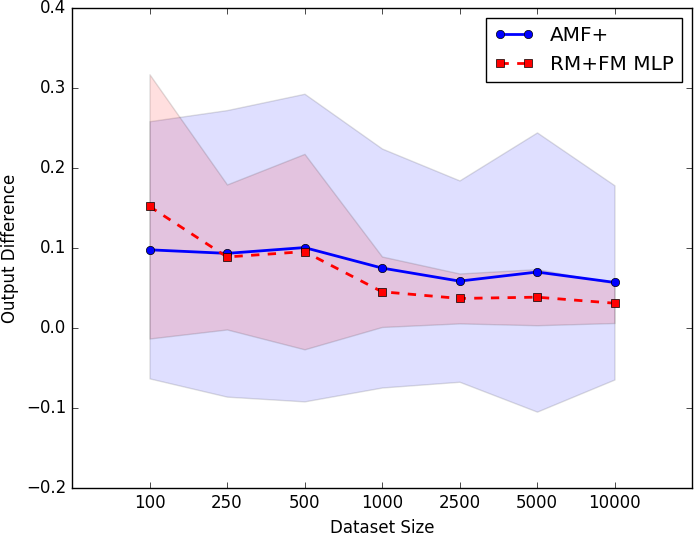}
\caption{Comparison of the performance of the RM$+$FM MLP neural network for the Civil Violence domain for varied dataset sizes (showing mean and standard deviation). Reducing the size of the training data increases the error in prediction (\textit{Output Difference} measured as Euclidean distance (see \cite{budhraja2018implementing}). The compared baseline indicates the performance of the AMF$^{+}$ framework for this domain in its current state as in \cite{budhraja2018improved}. The AMF$^{+}$ framework uses a total of $430$ instances of training data (see \cite{budhraja2019dissertation}) for the Civil Violence domain.}
\label{fig:usingNeuralNetworks:civil_violence}
\end{figure}

After establishing the use of neural networks as a potential scalable alternative to the AMF$^{+}$ framework, the next question to be answered is about the trade-off in data requirement. For this reason, the performance of the RM$+$FM network is evaluated at different dataset sizes (as done for \cite{brouwer2004feed} in \cite{budhraja2019dissertation}) for the Civil Violence domain. The use of $10$-fold cross-validation and test set size of $100$ are consistent with \cite{budhraja2019dissertation}. The results of this experiment are shown in Figure \ref{fig:usingNeuralNetworks:civil_violence}. Using the AMF$^{+}$ framework is observed to be more advantageous than using neural network architecture below a threshold of dataset size (approximately $1000$ in Figure \ref{fig:usingNeuralNetworks:civil_violence}). For larger datasets, the RM$+$FM network performs competitively and with significantly less variance. The large variance values observed for the AMF$^{+}$ framework indicate its susceptibility to outliers. This is also reflected by the wider distribution spread for the AMF$^{+}$ framework compared to the MLP network shown in Figure \ref{fig:usingNeuralNetworks:comparison}. The inference from Figure \ref{fig:usingNeuralNetworks:civil_violence} is therefore based on the trend in values rather than specific values themselves.

\subsection{Proposed One-to-Many Architecture}
\label{subsection:usingNeuralNetworks:experiments:proposedOnetomanyArchitecture}

Because the proposed MLP architecture is evaluated to be effective only for the Civil Violence ABM domain, the evaluation of the one-to-many variant of this architecture focuses on that domain.

\begin{figure*}[!t]
\centering
\subfloat[]{\includegraphics[width=0.33\textwidth]{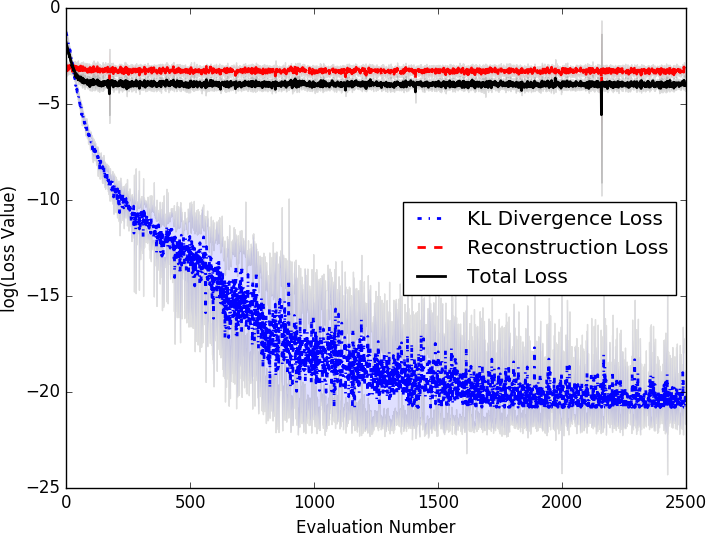}}
\subfloat[]{\includegraphics[width=0.33\textwidth]{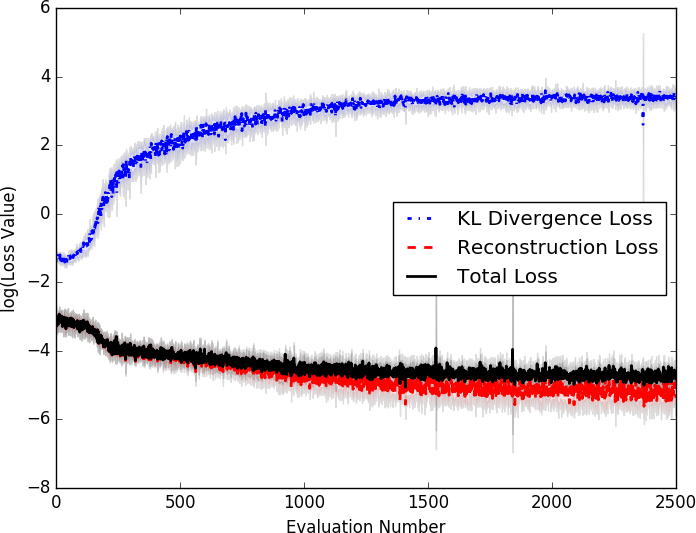}}
\subfloat[]{\includegraphics[width=0.33\textwidth]{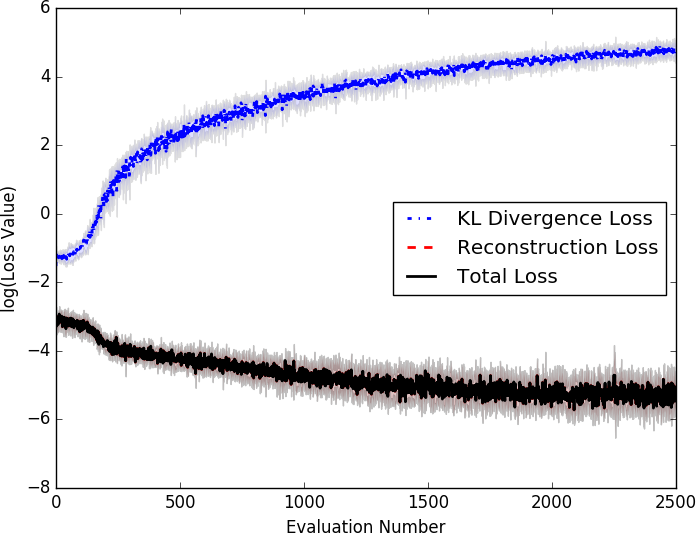}}
\caption{Mean and standard deviation values of KL divergence loss, reconstruction loss and total loss corresponding to the variational RM$+$FM network using $\alpha=0.5$, $\alpha=0.0001$ and $\alpha=0.0$ (($a$), ($b$) and ($c$) respectively). Values are averaged over $10$ cross-validation datasets for the Civil Violence ABM domain. Note that \textit{Reconstruction Loss} and \textit{Total Loss} are not distinguishable in ($c$) because of complete overlap caused by equality in value.}
\label{fig:usingNeuralNetworks:variational_loss}
\end{figure*}

Similar to Section \ref{subsection:usingNeuralNetworks:experiments:proposedMlpArchitecture}, RM$+$FM variational network MSE is observed when training the concatenated network. Experiments are summarized in Table \ref{tab:usingNeuralNetworks:neural_network_variational_mse}. To observe the effect of the use of KL divergence loss combined with reconstruction loss, we first observe a setting with $\alpha=0.5$. The performance of the RM$+$FM network shows significant degradation. It is therefore hypothesized that the KL divergence component of the loss shadows reconstruction loss. The loss is visualized in Figure \ref{fig:usingNeuralNetworks:variational_loss}, averaged over $10$ cross-validation datasets. Because of stability of convergence, the observations are limited to $2500$ evaluations of loss. Because the compared values may span orders of magnitude, performance is measured on a logarithmic scale. For computational convenience when using the logarithm, values which are evaluated to zero are replaced by the minimum non-zero value observed in the experiment. It is evident from Figure \ref{fig:usingNeuralNetworks:variational_loss} that KL divergence loss dominates reconstruction loss during early training iterations. The network therefore prioritizes the reduction of KL divergence loss. Once KL divergence loss is significantly lower than reconstruction loss, however, the optimizer does not proceed to reduce reconstruction loss. This is likely a local optima characteristic of the optimization surface. The network is then evaluated with a reduced contribution from KL divergence loss (using $\alpha=0.0001$). Prioritizing the optimization of reconstruction loss assists the optimizer in attaining improved network train and test MSE values, at the cost of significantly increased KL divergence error. Because of the observed improvement in MSE on decreasing $\alpha$, a final setting using $\alpha=0.0$ is evaluated. As expected based on the two previous results, the removal of the KL divergence loss component improves MSE performance of the network.

\begin{figure}
\centering
\includegraphics[width=0.45\textwidth]{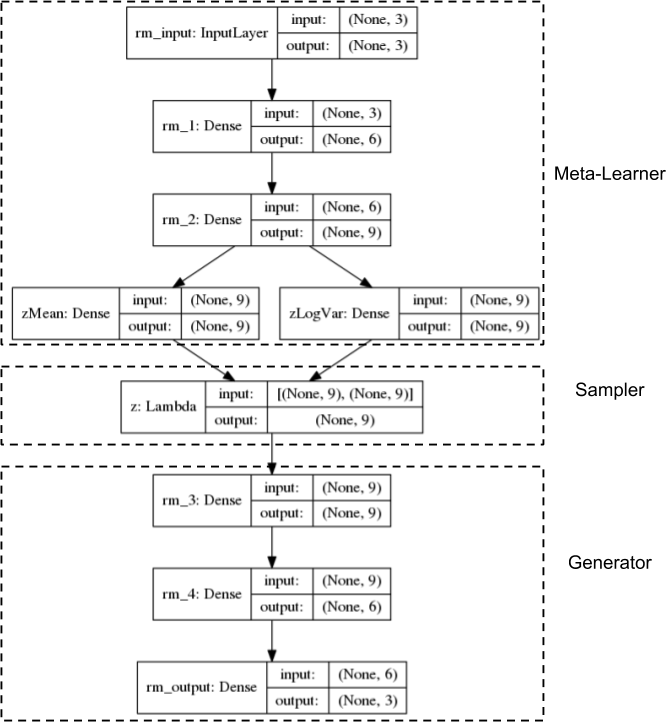}
\caption{Decomposition of the RM variational network into three logical components: the \textit{Meta-Learner} network, the \textit{Sampler} network and the \textit{Generator} network (top to bottom). The \textit{Meta-Learner} network learns to control Gaussian input to the \textit{Generator} network to produce ALP suggestions corresponding to input SLPs.}
\label{fig:usingNeuralNetworks:rm_parts}
\end{figure}

To understand the meaning of training the RM$+$FM variational network using $\alpha=0.0$, we first summarize the concept of meta-learning \cite{andrychowicz2016learning,finn2017model}. Meta-learning, or learning to learn, is the training of one learning model to learn to influence the training of a second learning model. This is used to explain the architecture of the RM component of the RM$+$FM variational network. In the case of the RM variational network, both these learning models are neural networks. We now revisit the RM network and divide it into three logical parts as shown in Figure \ref{fig:usingNeuralNetworks:rm_parts}. These parts are called the \textit{Meta-Learner} network, \textit{Sampler} network and \textit{Generator} network (in linear order). The role of the \textit{Generator} network is to produce an ALP suggestion based on the Gaussian signal from the \textit{Sampler} network. This Gaussian signal is controlled by the \textit{Meta-Learner} network. The \textit{Meta-Learner} network therefore learns to control Gaussian distributions to influence the learning of the \textit{Generator} network.

\begin{table}[!t]
\centering
\caption{Mean ($\mu$) and standard deviation ($\sigma$) for MSE computed for the concatenated RM$+$FM variational architecture for the Civil Violence ABM domain. After the completion of training, MSE is evaluated on the train and test data for a given cross-validation split.}
\begin{tabular}{c | c c | c c}
\toprule
\toprule
Alpha Value & \multicolumn{2}{c}{Train MSE} & \multicolumn{2}{c}{Test MSE $\mu$} \\
\cline{2-5}
{} & $\mu$ & $\sigma$ & $\mu$ & $\sigma$ \\
\midrule
0.5000 & 0.0380 & 0.0003 & 0.0381 & 0.0013 \\
0.0001 & 0.0012 & 0.0001 & 0.0012 & 0.0000 \\
0.0000 & \textbf{0.0003} & 0.0002 & \textbf{0.0003} & 0.0002 \\
\bottomrule
\bottomrule
\end{tabular}
\label{tab:usingNeuralNetworks:neural_network_variational_mse}
\end{table}

\begin{figure}
\centering
\includegraphics[width=0.45\textwidth]{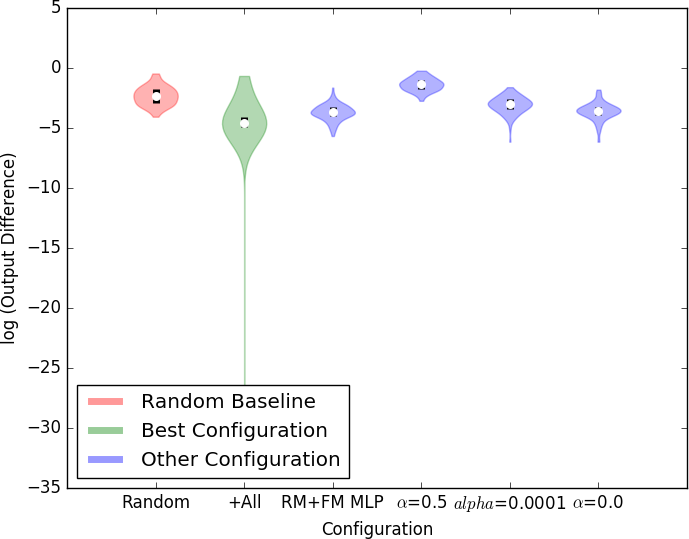}
\caption{Comparison of the performance of the performance of using them RM$+$FM variational network and the AMF$^{+}$ framework for the Civil Violence ABM. The variational network performs better with reduced values of $\alpha$. This Figure is a superset of Figure \ref{fig:usingNeuralNetworks:comparison}.}
\label{fig:usingNeuralNetworks:comparison_variational}
\end{figure}

Following Section \ref{subsection:usingNeuralNetworks:experiments:proposedMlpArchitecture}, the three configurations evaluated in Table \ref{tab:usingNeuralNetworks:neural_network_variational_mse} are then evaluated for their suggested ALPs i.e., the output at the end of the RM section of the network. Results for this evaluation are shown in Figure \ref{fig:usingNeuralNetworks:comparison_variational}. The median values of the logarithm of \textit{Output Difference} corresponding to $\alpha=0.5$, $\alpha=0.0001$ and $\alpha=0.0$ for the Civil Violence domain are $-1.3586$, $-2.9714$ and $-3.6011$ respectively. The values for the other methods shown in the figure are the same as Figure \ref{fig:usingNeuralNetworks:comparison}. Using $\alpha=0.5$ performs worse than the random baseline. This is improved by using $\alpha=0.0001$. Using $\alpha=0.0$ further improves the performance of this architecture and is competitive to the performance of the MLP network ($-3.7008$). Similar to the proposed MLP architecture, the proposed one-to-many architecture may therefore serve as a scalable alternative to the AMF$^{+}$ framework. This is with the additional constraint of produces multiple ALP points per SLP query, as opposed to the single ALP point produced by the MLP architecture.

\begin{figure*}[!t]
\centering
\subfloat[]{\includegraphics[width=0.33\textwidth]{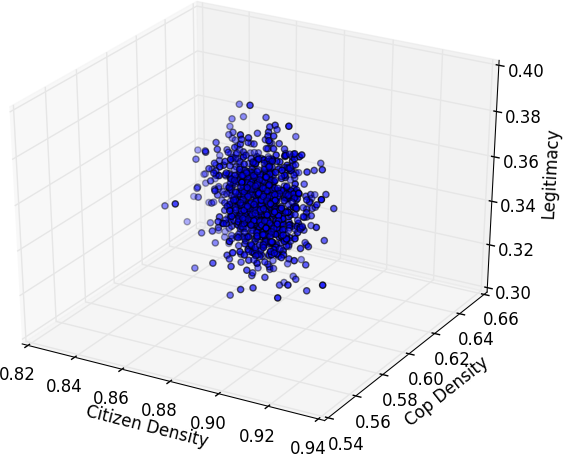}}
\subfloat[]{\includegraphics[width=0.33\textwidth]{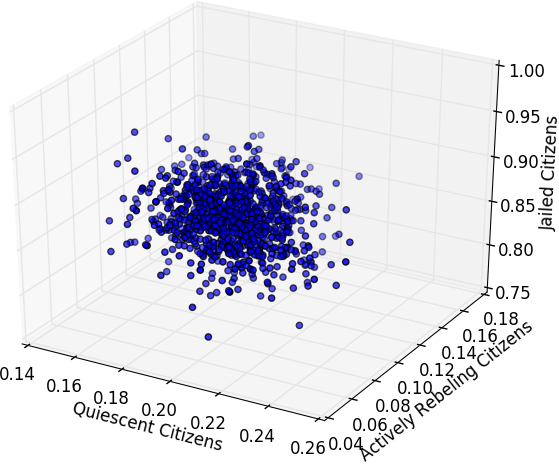}}
\subfloat[]{\includegraphics[width=0.33\textwidth]{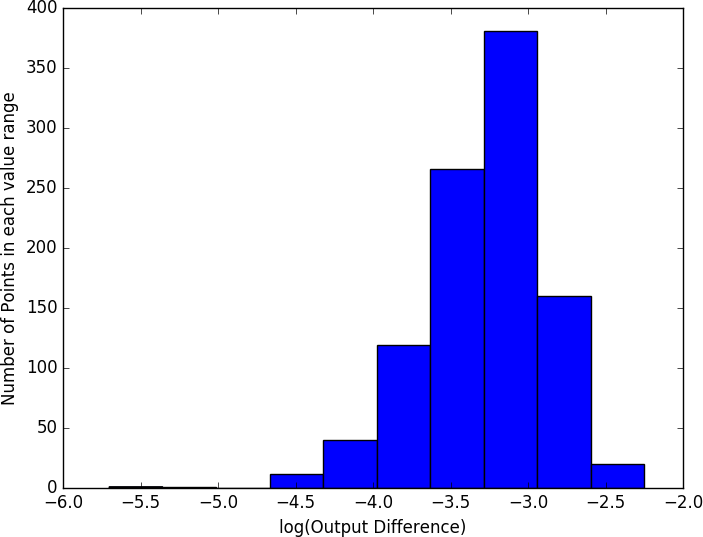}}
\caption{Distribution of points returned by the RM$+$FM variational network for the Civil Violence ABM domain using an arbitrary cross-validation test set query when trained on the given cross-validation training set. The points are shown in ($a$) ALP space and ($b$) corresponding SLP space (the output of the FM network). Corresponding \textit{Output Difference} values are summarized in ($c$). Note that all components of ALP and SLP vectors are normalized as in \cite{budhraja2016controlling,budhraja2017controlling,budhraja2019dissertation}.}
\label{fig:usingNeuralNetworks:architecture_civil_violence_variational_points}
\end{figure*}

For visual analysis, the points returned by the variational framework (using $\alpha=0.0$) are summarized in Figure \ref{fig:usingNeuralNetworks:architecture_civil_violence_variational_points}. Specifically, the network is trained using a cross-validation training set and then queried using an SLP configuration specified in the corresponding test set. The suggested ALP points are then used for ABM simulation to produce corresponding \textit{Output Difference} values. In SLP space, these points form a sphere proximal to the demonstration SLPs ($0.2142,0.0839,0.8883$) \cite{budhraja2017controlling,budhraja2019dissertation}. The spherical shape reflects the stability of the learned transformation between SLP and ALP spaces, by the network. In ALP space, these points are observed to form an ellipsoidal shape. The points shows a wider spread in ALP space than in SLP space. This reflects that points with minor perturbations in ALP parameters may still produce the same behavior for the Civil Violence ABM domain. This is not, however, an assumption that the network is based on. The network remains applicable to domains in which points that are significantly different in ALP space may produce the same output behavior. The values of \textit{Output Difference} observed are distributed similar to that observed in Figure \ref{fig:usingNeuralNetworks:comparison_variational}.

\section{Conclusion}
\label{section:usingNeuralNetworks:conclusion}

A framework for replicating abstract demonstrations (AMF$^{+}$) is discussed in \cite{budhraja2016controlling,budhraja2017controlling,budhraja2019dissertation}. Inherent from the AMF framework, the AMF$^{+}$ framework is limited in scalability, primarily with respect to the length of ALP vector \cite{miner2010dissertation}. As a scalable alternative, this work explores the use of neural networks to suggest ALPs for given SLPs. The proposed RM$+$FM MLP and variational neural network architectures are competitive with the performance of the AMF$^{+}$ framework beyond a threshold of data availability (when the FM network excels at approximating the ABM). Training the RM$+$FM network also results in significantly improved network training (see Table \ref{tab:usingNeuralNetworks:neural_network_mse_mean_sigma}). This shows a method for potential improvement in performance for existing works that only employ a direct RM mapping (such as those discussed in Section \ref{section:relatedWork}, with the exception of the use of feedback \cite{almusawi2016new,jordan1992forward}).

The proposition of a neural network for these domains may further be explored for time series data, images and video by integrating Recurrent Neural Network (RNN) layers, Convolutional Neural Network (CNN) layers and Recurrent Convolutional Neural Network (RCNN) layers respectively. The network architecture for one-to-many mapping currently uses variational layers. Alternatively, the network may be trained to learn a distribution surface using Mixture Density Network (MDN) layers \cite{bishop1994mixture}. To avoid manual specification of the number of contributors to the probability surface (such as the number of Gaussian distributions composited), we may use regularization. This can be done by starting with a large number of composite factors and then using regularization to reduce redundant contributions.

\bibliographystyle{IEEEtran}
\bibliography{IEEEabrv,swarm-lfd}

\end{document}